\documentclass[a4paper, 10pt, conference, multi]{ieeeconf}      % Use this line for a4
                                                          
\usepackage{FG2020}
\usepackage{subcaption}
\usepackage{graphicx}
\usepackage[capitalise]{cleveref}
\usepackage{url}

\FGfinalcopy % *** Uncomment this line for the final submission

\IEEEoverridecommandlockouts                              % This command is only
\def\FGPaperID{3} % *** Enter the FG2020 Paper ID here

\title{
\LARGE \bf Complement Face Forensic Detection and Localization with Facial Landmarks
}

\author{\parbox{16cm}{\centering
    {\large Kritaphat Songsri-in$^1$ and Stefanos Zafeiriou$^{1,2}$}\\
    {\normalsize
    $^1$ Department of Computing, Imperial College London, UK\\
    $^2$ Center for Machine Vision and Signal Analysis, University of Oulu, Finland}}\\
    \{\textit{kritaphat.songsri-in11, s.zafeiriou\}@imperial.ac.uk}
    \thanks{This work was not supported by any organization}
}

\begin{document}

\ifFGfinal
\thispagestyle{empty}
\pagestyle{empty}
\else
\author{Anonymous FG2020 submission\\ Paper ID \FGPaperID \\}
\pagestyle{plain}
\fi
\maketitle

%%%%%%%%%%%%%%%%%%%%%%%%%%%%%%%%%%%%%%%%%%%%%%%%%%%%%%%%%%%%%%%%%%%%%%%%%%%%%%%%
\begin{abstract}
Recently, Generative Adversarial Networks (GANs) and image manipulating methods are becoming more powerful and can produce highly realistic face images beyond human recognition which have raised significant concerns regarding the authenticity of digital media. Although there have been some prior works that tackle face forensic classification problem, it is not trivial to estimate edited locations from classification predictions. In this paper, we propose, to the best of our knowledge, the first rigorous face forensic localization dataset, which consists of genuine, generated, and manipulated face images. In particular, the pristine parts contain face images from CelebA and FFHQ datasets. The fake images are generated from various GANs methods, namely DCGANs, LSGANs, BEGANs, WGAN-GP, ProGANs, and StyleGANs. Lastly, the edited subset is generated from StarGAN and SEFCGAN based on free-form masks. In total, the dataset contains about 1.3 million facial images labelled with corresponding binary masks.

Based on the proposed dataset, we demonstrated that explicit adding facial landmarks information in addition to input images improves the performance. 
In addition, our proposed method consists of two branches and can coherently predict face forensic detection and localization to outperform the previous state-of-the-art techniques on the newly proposed dataset as well as the faceforecsic++ dataset especially on low-quality videos.
\end{abstract}

%%%%%%%%%%%%%%%%%%%%%%%%%%%%%%%%%%%%%%%%%%%%%%%%%%%%%%%%%%%%%%%%%%%%%%%%%%%%%%%%
\section{INTRODUCTION}
Face images and videos have always been at the focus of the machine learning and computer vision community with various supervised learning problems such as face detection, face alignment, face recognition, \emph{etc}. Their applications span from surveillance system \cite{Zafeiriou:2015:SFD:2805319.2805395, journals/cai/LuZY03}, autofocus on digital camera \cite{violaobject, faceincamera}, and face verification \cite{facenet, vggface, lightcnn}. 

Meanwhile, since \cite{gan} introduced Generative Adversarial Networks (GANs) in 2014 as a core framework for a generative model with deep learning, many works \cite{dcgan, lsgan, began, wgangp, sagan}, have gradually improved the method in term of training stability and image quality. Notably, \cite{progan, stylegan} proposed revolutionary architectures and training procedure to generate hyper-realistic face images at high-resolution. Their results have achieved an unprecedented level of details that are hardly distinguishable by humans. Additionally, the qualities of automatic face editing methods have also significantly been improved. For instance, \cite{stargan} proposed to edit face images based on discrete target attributes; \cite{scfegan} can perform image completion based on user sketch and target colours; \cite{face2face} remarkably present a method for real-time facial reenactment based on RGB images. With the advance in both generative models and image manipulation methods, it is almost impossible for humans to easily detect them making digital forensic regarding face images and videos indispensable.

\begin{figure}
    \centering
    \includegraphics[width=\linewidth]{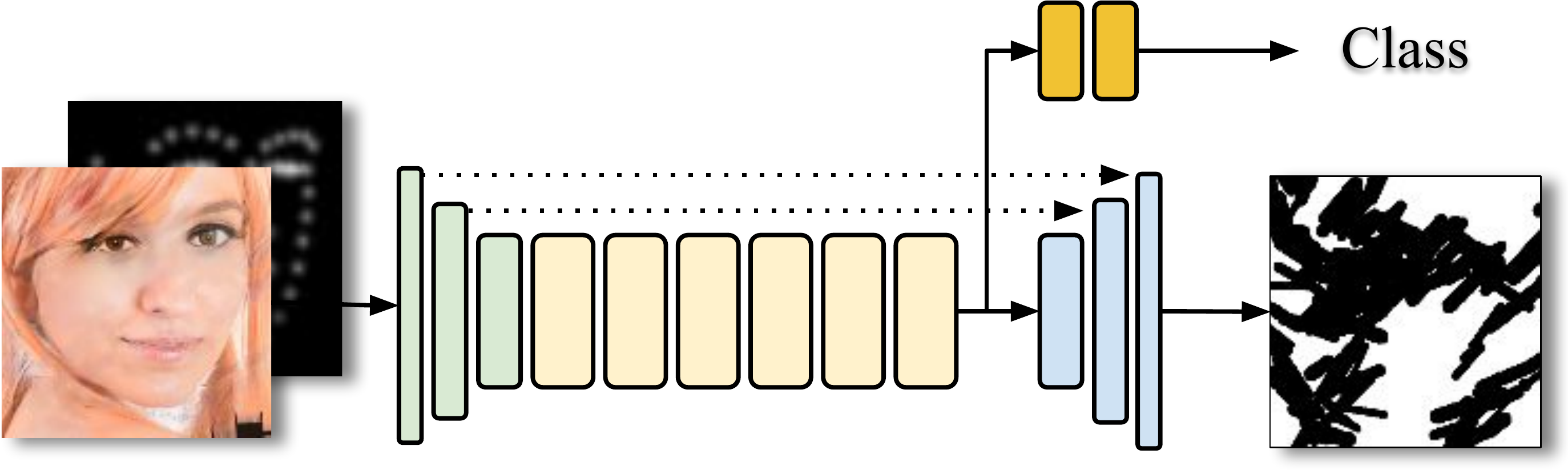}
    \caption{We introduce a large scale face forensic localization dataset and propose a DCNN that utilize spatial facial landmarks information and combine complementary classification and localization predictions.}
    \label{fig:firstfig}
\end{figure}

There are some preliminary works such as \cite{detrctfakefacewithlandmarks, detectfakeface} that try to identify real facial images from generated images. Similarly, the works in \cite{mesonet, faceforensic++} proposed dataset based on image editing methods in \cite{deepfakes, faceswap, face2face} and proposed DCNNs to solve a face forensic detection problem. Nevertheless, none of these methods combines the problem of distinguishing real images from fake images as well as detecting edited images altogether. Besides, most of their datasets are usually only contain class labels but not corresponding binary masks. 

In this paper, we propose to directly solve face forensic localization by introducing a new dataset that consists of pristine images, generated images, and partially edited images to develop a model that can jointly solve face forensic detection and localization as depicted in \cref{fig:firstfig}. In summary, the contributions of our work are 
\begin{itemize}
    \item We proposed a large scale face forensic localization dataset labelled with corresponding binary masks. The dataset consists of about 1.3 million facial images which contain real image, generated images, and partially edited images.
    \item We utilize a spatial feature from facial landmarks in order to improve face forensic detection and localization.
    \item Our novel architecture is based on an XceptionNet to exploit transfer learning and is adjusted to output both classification and localization predictions.
    \item When the classification and localization predictions are holistically combined during training, the performance of face forensic localization can be further improved.
\end{itemize}

% %%%%%%%%%%%%%%%%%%%%%%%%%%%%%%%%%%%%%%%%%%%%%%%%%%%%%%%%%%%%%%%%%%%%%%%%%%%%%%%%

\section{RELATED WORKS}

\subsection{Generative Adversarial Networks (GANs)} 
Generative Adversarial Networks (GANs) \cite{gan} are generative models that consist of two competing networks: a generator and a discriminator. The discriminator objective is to distinguish generated samples from real samples. On the contrary, the generator's goal is to produce realistic samples in order to fool the discriminator. The original framework in \cite{gan} was initially developed for image generation but are widely adopted to other tasks such as conditional image generation \cite{cvae} and image-to-image translations \cite{pix2pix, CycleGAN2017, Recycle-GAN, stargan}. Due to its popularity, there are many variations of the frameworks that extend the method and improve upon image quality as well as the stability during training by adjusting networks' architecture \cite{dcgan, began, stylegan}, loss functions \cite{wgan, began}, or training procedure \cite{ttur, progan}, \emph{etc}.

\subsection{Face Manipulation Methods}
\subsubsection{StarGAN}
The method in \cite{stargan} can translate images between domains without having target ground truths through the use of cycle-consistency loss when the inverted image domain translation is performed similarly to \cite{pix2pix, CycleGAN2017, Recycle-GAN}. When translating between two domains, the method in \cite{pix2pix, CycleGAN2017, Recycle-GAN} rely on different generators. However, rather than having a different generator for every two possible domains, StarGAN proposed to tackles the problem differently. In particular, they proposed to solve multiple domains translation by fusing target domain attributes with a given image. Concatenating the target attributes and the input image channel-wise allows a single generator to learn shareable feature between similar domains such as changing skin colours, hairstyles, and facial emotions.
\subsubsection{SC-FEGAN}
SC-FEGAN \cite{scfegan} is a face image completion method that takes incomplete images and sketches at the missing areas as inputs and output compatible realistic results. The method is based on GANs and gated convolutional layers that provide a learnable spatial feature selection mechanism. During training, the method imitates user incomplete inputs with randomly generated free-form masks and corresponding missing ground truth sketch. Their method can produce realistic face images with arbitrary masks and sketches at high resolutions $512 \times 512$.

\subsection{Face Forensic Detections}

\subsubsection{Generated Face Images Detections}
Recently, newly proposed GANs \cite{progan, stylegan} can generate an unprecedented level of face image quality at a larger resolution which in term increase concerns regarding fake or generated face images classification. There are some primary works that tackle this problem. \cite{detectfakeface} proposed to classify fake images with a DCNN that not only learn to classify images with a cross-entropy loss but also utilize a contrastive loss that can project images with the same class closer together in a latent space based on a Euclidean distance. On the other spectrum, \cite{detrctfakefacewithlandmarks} proposed to exploit inconsistent configurations of facial landmarks due to weak global constraints of generated images. The extracted normalized face landmarks are used as a feature for a Support Vector Machine (SVM) classifier and achieve competitive results with DCNN based methods.
Interestingly, some of the methods that modify discriminators to better classify generated images in GANs literature could shed some light on the fake image detection problem as well. For example, \cite{sagan} introduced a self-attention module which allows the discriminator to classify images based on attention-driven, long-range dependency. \cite{began} also proposed a different idea for the auto-encoder based discriminator to differentiate images through image reconstruction loss rather than unstable standard adversarial losses in \cite{gan}.

\subsubsection{Manipulated Face Forensics Detections}
\cite{mesonet, faceforensic++} both introduced face forensic detection datasets based on automated face image manipulations: DeepFake \cite{deepfakes} and Face2Face \cite{face2face}. \cite{mesonet} proposed a shallow network to capture the mesoscopic properties of the images, called MesoInception-4. The network contains four modules: two inception modules \cite{inception} followed by two standard convolution modules where a batch normalization layer and a max-pooling layer are inserted after each module. Lastly, two layers of fully connected layers are used to output the predictions. They also use a mean squared error instead of a cross-entropy loss. \cite{faceforensic++} added a face manipulating method, FaceSwap \cite{faceswap} to the dataset and instead utilized transfer learning by fine-tuning a XceptionNet \cite{Xception} pre-trained on the ImageNet dataset \cite{ILSVRC15}. Although, manipulated videos in \cite{faceforensic++} are also labelled with ground truth masks which can be used for face forensic localization, they did not report the numbers on localization problem on their proposed dataset.

\section{FACE FORENSICS LOCALIZATION DATASET}
\begin{figure*}
\includegraphics[width=\textwidth]{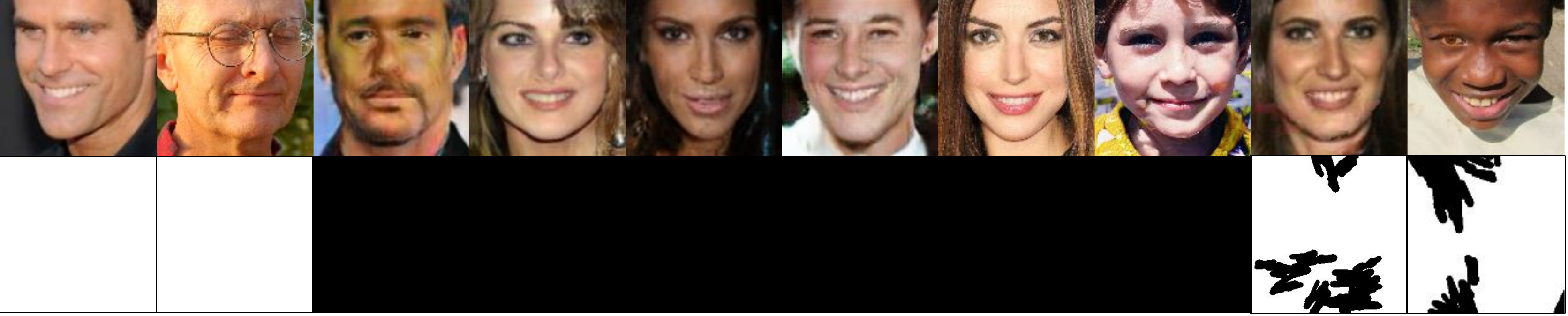}
\captionsetup[subfigure]{aboveskip=-1pt,belowskip=-1pt}
\begin{minipage}[t]{.1\linewidth}
\vspace{-0.8\baselineskip}
\subcaption{\scriptsize CelebA}
\end{minipage}
\begin{minipage}[t]{.1\linewidth}
\vspace{-0.8\baselineskip}
\subcaption{\scriptsize FFHQ}
\end{minipage}\begin{minipage}[t]{.1\linewidth}
\vspace{-0.8\baselineskip}
\subcaption{\scriptsize DCGAN}
\end{minipage}\begin{minipage}[t]{.1\linewidth}
\vspace{-0.8\baselineskip}
\subcaption{\scriptsize LSGAN}
\end{minipage}\begin{minipage}[t]{.1\linewidth}
\vspace{-0.8\baselineskip}
\subcaption{\scriptsize BEGAN}
\end{minipage}\begin{minipage}[t]{.1\linewidth}
\vspace{-0.8\baselineskip}
\subcaption{\scriptsize WGANGP}
\end{minipage}\begin{minipage}[t]{.1\linewidth}
\vspace{-0.8\baselineskip}
\subcaption{\scriptsize ProGAN}
\end{minipage}\begin{minipage}[t]{.1\linewidth}
\vspace{-0.8\baselineskip}
\subcaption{\scriptsize StyleGAN}
\end{minipage}\begin{minipage}[t]{.1\linewidth}
\vspace{-0.8\baselineskip}
\subcaption{\scriptsize StarGAN}\label{fig:stargan}
\end{minipage}\begin{minipage}[t]{.1\linewidth}
\vspace{-0.8\baselineskip}
\subcaption{\scriptsize SCFEGAN}\label{fig:scfegan}
\end{minipage}
\caption{\label{fig:samples} Sample images of the proposed dataset which contain real, generated, and partially edited face images. The second row shows corresponding binary map where white pixels represent pristine locations and vise versa.}
\end{figure*}

One of the contributions of this paper is proposing a new face forensic localization dataset containing real images, generated images, and partially edited images. The labels for real and generated images are binary images whose pixels values are completely 1 and 0, respectively. However, the partially edited images are created depending on the randomly generated free form masks \cite{gatedconv, scfegan}. Part of the images taken from real images has labels 1 while the pixels taken from the counterfeit part are labelled with 0. Sample images and corresponding binary masks of each class is shown in \cref{fig:samples}.

\subsection{Real Face Images}
Pristine face images of the proposed dataset contains images from Large-scale CelebFaces Attributes (CelebA) \cite{celeba} and Flickr-Faces-High-Quality (FFHQ) \cite{stylegan} datasets. The CelebA dataset is a large-scale face dataset which contains 202,599 celebrity images at $178 \times 218$ pixels. Each image is annotated with 40 attributes as well as facial landmarks. The dataset covers large pose variations, diverse ethnicity, and different background. The FFHQ dataset consists of 70,000 high-quality PNG face images at $1024\times1024$ resolution. The dataset contains considerable variation in terms of age, ethnicity and background but has good coverage of accessories such as eyeglasses, sunglasses, and hats. In total, the proposed dataset consists of 272,599 pristine face images.

\subsection{Generated Face Images}
In this work, we add samples generated from extensive versions of GANs that are representative in term of networks' architectures, loss functions, and training procedures. This includes DCGANs \cite{dcgan}, LSGANs \cite{lsgan}, BEGANs \cite{began}, WGANP-GP \cite{wgangp}, ProGANs \cite{progan}, and StyleGANs \cite{stylegan}. Apart from ProGANs and StyleGANs where we take their 100,000 generated face images at $1024 \times 1024$, each of these GANs is trained to generate 100,000 face images at $128 \times 128$ resolution. As a result, we have 600,000 generated images.

\subsection{Partially Edited Face Images}
We generate partially edited face images based on randomly generated free form mask. The counterfeit part of the images is simulated with two methods StarGANs and SC-FEGAN. In total, the partially edited images consist of 202,599 images from StarGAN and 272,599 images from SC-FEGAN.
\subsubsection{StarGAN}
We used the official implementation \cite{stargan} trained at $256 \times 256$ on the CelebA dataset to create counterfeit images based on ground-truth attributes \emph{i.e.} recreate face images with the same attributes rather than changing its attributes.
\subsubsection{SC-FEGAN}
We used the official implementation \cite{scfegan} trained on the CELEBA-HD dataset \cite{progan}. We generated corresponding counterfeit images for both CelebA and FFHQ datasets.

In summary, the whole dataset contains 1,347,797 face images. We split the images into two folds: the first 80\% of each class is used for training and validation while the last 20\% of each class is used for testing.

\subsection{Pre-processing}
In order to alleviate the problem of face forensic localization, the dataset is assumed to contain a single face image at the centre of the image. To remove bias between images' classes and images' resolutions, each image is first randomly resized to fall between 0.8 to 1.2 scale of the target $128 \times 128$ resolutions. We also align face images base on their facial landmarks. Although CelebA and FFHQ datasets are annotated with sparse 5 points facial landmarks extracted from dlib \cite{dlib}, we applied state-of-the-arts face detection and alignment method in \cite{jiankangfg2018} to retrieve denser commonly used 68 points facial landmarks \cite{sagonas_iccv_300w_2013, 300VW} for all images of the proposed dataset. With the extracted facial landmarks, each image is aligned to canonical face landmarks using similarity alignment. To avoid zero-padding, each image is then padded with mirror image and cropped at resolution $128 \times 128$.

\section{MODEL}
\begin{figure*}
    \centering
    \includegraphics[width=\linewidth]{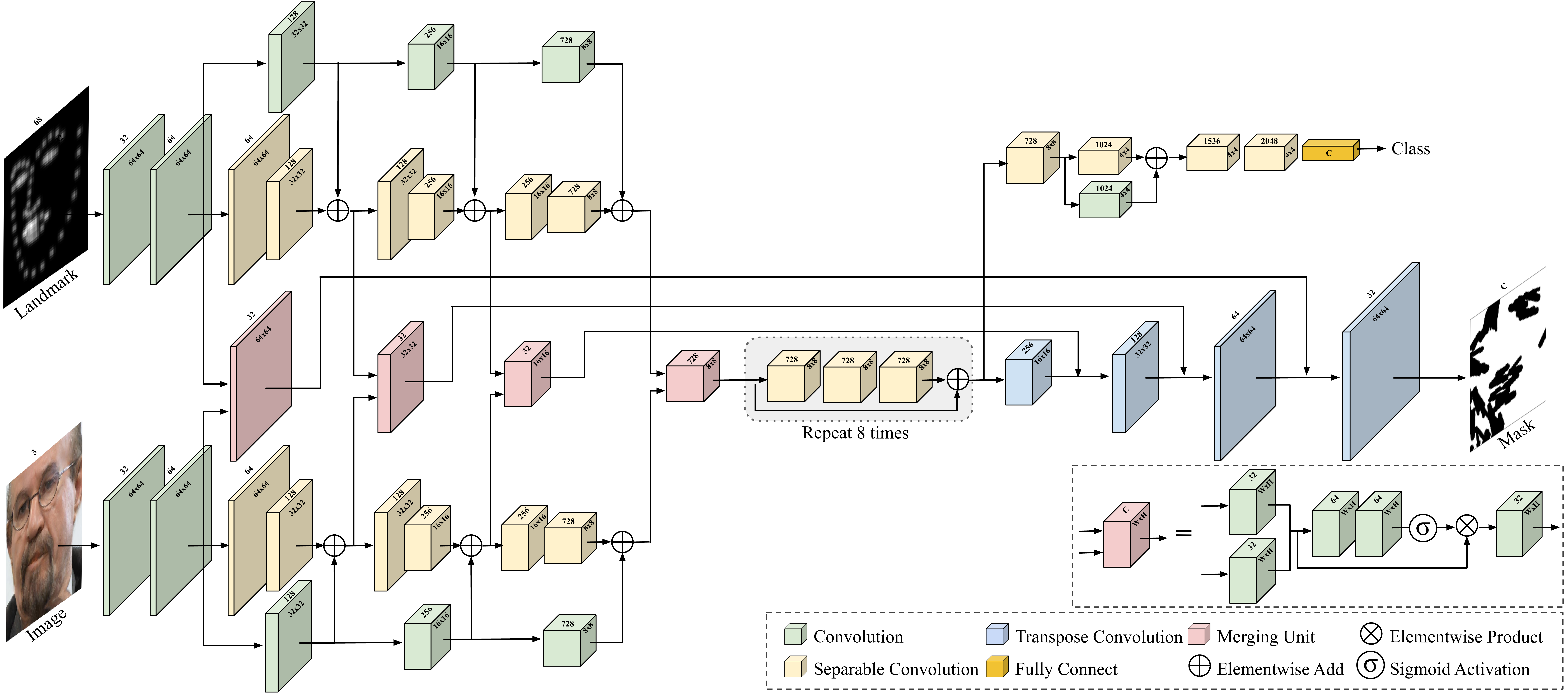}
    \caption{Our network architecture consists of four parts: an encoder, residual modules, a class prediction branch, and a binary mask prediction branch.}
    \label{fig:network}
\end{figure*}

To fully exploit the complement between face forensic classification and localization on the aligned face images $I$, we consider a network that can output two branches: one for classifications $O_{class}(I)$ and the other for localizations $O_{mask}(I)$. For a dataset with $C$ classes, the classification branch output $C$ logits to be passed through a soft-max layer in order to make a prediction, $P_{class}(I)=\textrm{softmax}(O_{class}(I))$ while the localization branch output a prediction with the same resolution as the given image. To abbreviate notations, we will omit input face image $I$ from our equations. The entire architecture is depicted in \cref{fig:network}.

\subsection{Back Bone Architecture}
We adopt an XceptionNet \cite{Xception} pre-trained on the ImageNet dataset as a backbone network. We left the original class prediction branch intact and added a mask prediction branch after a repeating middle flow module. In particular, the newly introduced mask prediction branch consist of 3 transposed convolution layers each concatenated with a skip connection from \texttt{Conv\_2}, \texttt{Residual\_1}, and \texttt{Residual\_2} layers respectively. Lastly, a convolution layer is used to adjust the output's channel.

\subsection{Combining Classification and Localization branches}
\subsubsection{Shared single binary mask}
The localization branch for this variation is shared among all classes \emph{i.e.} the output is a binary mask corresponding to a prediction of a given image. In this scenario, the class prediction branch and the localization branch are independent, and each can be trained for their own corresponding losses. For a given image with shape $[W, H, 3]$, the localization branch output of this variation has shape,  $|O_{class}|=[W, H, 1]$, and the prediction can be defined through a sigmoid function for each pixel as:
\begin{equation}
    P_{mask} = \textrm{sigmoid}(O_{mask})
    \label{eq:hard}
\end{equation}
\subsubsection{Independent binary mask for each class}
In this case, we try to output $C$ binary mask, one for each class. The localization outputs are now binary masks whose shape is $[W, H, C]$. During training, only a mask at the ground truth class is optimized. At inferencing, a binary mask with the highest probability based on the class prediction branch is chosen as a localization output. Formally, the binary mask prediction $P_{mask}$ is defined as
\begin{equation}
    P_{mask} = \textrm{sigmoid}(O_{mask}(c)), c = \textrm{argmax}(P_{class})
    \label{eq:soft}
\end{equation}
In this variation, the localization branch pushes the class prediction responsibility to a class prediction branch and can exploit any bias associated with each class \emph{e.g.}. real and fake are always binary mask with values 1 and 0 respectively. 
\subsubsection{Soft independent binary mask for each class}
The previous method only makes localization prediction based on a single binary mask whose class prediction is highest, which can be thought of as a hard version when making the decision. We proposed to use a soft version where all binary mask contributes to the final prediction by proportionally averaging the logit outputs from the class prediction branch before the sigmoid function formally described as:
\begin{equation}
    P_{mask} = \textrm{sigmoid}(\frac{1}{C}\sum_{c \in C} P_{class}(c) * O_{mask}(c))
\end{equation}
In the case that a class prediction is easy ($O_{class}(c)$ of the correct class is high), this version also has the benefit of reusing information from the class prediction branch similar to \ref{eq:hard}. However, in the case where the decision is unclear, this version can alleviate the localization output rather than harshly forced to choose a particular class's binary mask.
\subsection{Adding Facial Landmarks Information}
Facial landmarks can be used as an important spatial feature to detect manipulated face images as demonstrated by \cite{detrctfakefacewithlandmarks}.
However, rather than directly feeding the 2D landmarks as an extra 136 feature to the network, we incorporate them into the network as binary-maps with 68 channels. At each channel, only a pixel at the corresponding landmark position is label 1 and 0 otherwise. 
For a newly trained network, this landmark image can be channel-wisely concatenated with the image and together are used as an input to the network directly. However, since our proposed method utilizes networks' weights from a pre-trained network, we instead introduce a mirror head for the landmark image with the identical structure as the image head. The skip connections to the localization branch are now fused based on the activations of each branch with Self-Supervised Model Adaptation (SSMA) \cite{ssma}. The complete detail of the network is shown in \cref{fig:network}.
\subsection{Model Learning}
Since all of our variations output the predictions with the same shapes ($|P_{class}| = [C], |P_{mask}| = [W, H, 1]$), a cross-entropy loss function is used for both optimizing class and localization branches. In particular, the loss on mask prediction is computed by averaging the cross-entropy between each pixel of the predicted mask and the ground-truth mask $\tilde{P}_{mask}$. Let a cross-entropy between two distribution be $X$, and the ground-truth class label be $\tilde{P}_{class}$ then
\begin{equation}
    L = X(P_{class}, \tilde{P}_{class}) +  \lambda(\frac{1}{W*H}\sum_{i\in W, j\in H} X(P^{ij}_{mask}, \tilde{P}^{ij}))
\end{equation}
For each model that output two branches, we set $\lambda =100$.
The networks are trained with adaptive moment estimation optimizer (ADAM) \cite{adam} using parameters: $\alpha=0.0001$, $\beta_1=0.9$, and $\beta_2=0.999$ with batch size $64$. During training, we tackle the imbalanced samples problem by randomly sampling images from each class with equal probability.
\section{EXPERIMENTS}
To demonstrate the power of our proposed methods, we report face forensic detections, classifications, and localization results on the proposed dataset and FaceForecsic++ dataset against other state-of-the-arts. 
\subsection{Baseline Methods}
For face forensic detection and classification, we compare our proposed methods with the state-of-the-art methods in \cite{mesonet, faceforensic++}. For face forensic localization, we proposed two methods as baselines.
\subsubsection{Encoder-decoder} 
We use a simple encoder-decoder architecture proposed in \cite{stargan}. The original network was used for image-to-image translation task, and it can output a prediction mask at the same resolution as the given image.
\subsubsection{XceptionNet}
Similar to \cite{faceforensic++} where the pre-trained Xception is fine-tuned for face forensic detection, we report the naively pre-trained XceptionNet adjusted to output localization mask. Specifically, this baseline is our method without utilizing classification prediction or landmarks information.

\subsection{Proposed dataset}
\subsubsection{Face Forensic Detections}
In order to fully compare the methods, they are trained to predict 10 classes, each representing the source of the given image. We then report the results on three settings where the image' source can be regarded as real vs fake, real vs fake vs edited, or directly the image source.
The accuracy values are shown in \cref{table:ourdataset_binary}, \cref{table:ourdataset_soft_classify}, and \cref{table:ourdataset_hard_classify} respectively. For each of our proposed method, we also report binary classification according to a mask prediction in which a predicted mask with at least one edited pixel is considered as a fake image (labelled with ``Mask''). From the tables, our proposed method with soft version outperforms other state-of-the-arts in term of face forensic detection and face forensic type classification reaching 99.25\% and 99.16\% respectively. On the other hand, our share version performs best for image source classification at 98.85\%. Comparing our methods, we see a slight drop in performance when a hard version is used. This may be because although the mask prediction rely on the class prediction, it was not fully incorporated during training unlike our soft version.
\begin{table*}[]
\centering
\begin{tabular}{c|c|c|c|c|c|c|c|c|c|c|c|}
\cline{2-12}
 & CelebA & FFHQ & DCGAN & LSGAN & BEGAN & WGAN & ProGAN & StyleGAN & StarGAN & SC-FEGAN & Total \\ \hline
\multicolumn{1}{|l|}{MesoNet} &\textbf{98.86}  &86.09  &\textbf{100}  &\textbf{100}  &\textbf{100}  &99.80  &99.91  &97.46  &91.70  &91.86 &96.00  \\ \hline
\multicolumn{1}{|l|}{Encoder-Decoder} &98.09 & 98.66 &    99.00 & \textbf{100}    & 99.00    &98.99 & 98.91 & 98.67 & 98.47 & 97.11 &98.43 \\ \hline
\multicolumn{1}{|l|}{Xception} &96.28  &99.71  &\textbf{100}  &\textbf{100}  &\textbf{100}  &\textbf{100}  &99.76  &97.32  &98.75  &98.52  &98.72  \\ \hline \hline
\multicolumn{1}{|l|}{Share} &96.45  &\textbf{99.99}  &\textbf{100}  &\textbf{100}  &\textbf{100}  &\textbf{100}  &99.93  &99.78  &99.65  &98.53  &99.09  \\ \hline
\multicolumn{1}{|l|}{Share-mask} &97.91  &99.92  &\textbf{100}  &\textbf{100}  &\textbf{100}  &\textbf{100}  &\textbf{99.98}  &99.88  &\textbf{99.84} &97.86  &99.21  \\ \hline
\multicolumn{1}{|l|}{Hard} &96.01  &90.88  &99.98  &\textbf{100}  &\textbf{100}  &\textbf{100}  &99.90  &\textbf{99.99}  &98.81  &\textbf{98.76}  &98.49  \\ \hline
\multicolumn{1}{|l|}{Hard-mask} &98.80  &90.90  &99.98  &\textbf{100}  &99.99  &100  &99.90  &\textbf{99.99}  &99.35  &97.88  &98.81  \\ \hline
\multicolumn{1}{|l|}{Soft} &97.75  &99.89  &\textbf{100}  &\textbf{100}  &\textbf{100}  &\textbf{100}  &99.74  &99.18  &99.53  &98.46  &99.20  \\ \hline
\multicolumn{1}{|l|}{Soft-mask} &98.72  &99.89  &\textbf{100}  &\textbf{100}  &\textbf{100}  &\textbf{100}  &99.89  &99.80  &99.69  &97.62  &\textbf{99.25}  \\ \hline
\end{tabular}
\caption{Face forensic binary detection (real vs. fake) accuracy reported on the proposed dataset.}
\label{table:ourdataset_binary}
\end{table*}

\begin{table*}[]
\centering
\begin{tabular}{c|c|c|c|c|c|c|c|c|c|c|c|}
\cline{2-12}
 & CelebA & FFHQ & DCGAN & LSGAN & BEGAN & WGAN & ProGAN & StyleGAN & StarGAN & SC-FEGAN & Total \\ \hline
\multicolumn{1}{|l|}{MesoNet} &\textbf{98.86}  &86.09  &99.98  &99.94  &99.99  &99.56  &99.84  &97.31  &91.63  &90.89  &95.75  \\ \hline
\multicolumn{1}{|l|}{Encoder-Decoder} &98.09 & 98.66 &    99.00 & \textbf{100}    & 99.00    &98.99 & 98.91 & 98.67 & 98.46 & 97.07 &98.42 \\ \hline
\multicolumn{1}{|l|}{XceptionNet} &96.28  &99.71  &\textbf{100}  &\textbf{100}  &\textbf{100}  &\textbf{100}  &99.74  &97.30  &98.56  &98.37  &98.66  \\ \hline \hline
\multicolumn{1}{|l|}{Share} &96.45  &\textbf{99.99}  &\textbf{100}  &\textbf{100}  &\textbf{100}  &\textbf{100}  &\textbf{99.92}  &99.78  &\textbf{99.64}  &98.43  &99.07  \\ \hline
\multicolumn{1}{|l|}{Hard} &96.01  &90.88  &99.96  &\textbf{100}  &99.92  &\textbf{100}  &99.88  &\textbf{99.99}  &98.80  &\textbf{98.63}  &98.45  \\ \hline
\multicolumn{1}{|l|}{Soft} &97.75  &99.89  &\textbf{100}  &\textbf{100}  &\textbf{100}  &\textbf{100}  &99.68  &99.01  &99.53  &98.38  &\textbf{99.16}  \\ \hline
\end{tabular}
\caption{Face forensic type classification (real vs. fake vs. edited) accuracy reported on the proposed dataset.}
\label{table:ourdataset_soft_classify}
\end{table*}

\begin{table*}[]
\centering
\begin{tabular}{c|c|c|c|c|c|c|c|c|c|c|c|}
\cline{2-12}
 & CelebA & FFHQ & DCGAN & LSGAN & BEGAN & WGAN & ProGAN & StyleGAN & StarGAN & SC-FEGAN & Total \\ \hline
\multicolumn{1}{|l|}{MesoNet} &\textbf{98.86}  &86.09  &99.85  &99.66  &99.94  &99.12  &99.78  &97.29  &89.16  &90.49  &95.23  \\ \hline
\multicolumn{1}{|l|}{Encoder-Decoder} & 98.09    & 98.66 & 98.96    & 99.90 &98.96 & 98.98 & 98.88 & 98.66 & 98.45 & 97.06 & 98.40  \\ \hline
\multicolumn{1}{|l|}{XceptionNet} &96.28  &99.71  &\textbf{99.96}  &\textbf{99.99}  &\textbf{99.96}  &99.99  &99.74  &97.30  &98.35  &98.25  &98.60  \\ \hline \hline
\multicolumn{1}{|l|}{Share} &96.45  &\textbf{99.99}  &99.86  &98.64  &99.86  &99.94  &99.84  &99.78  &\textbf{99.06}  &98.42  &\textbf{98.85}  \\ \hline
\multicolumn{1}{|l|}{Hard} &96.01  &90.88  &99.66  &99.94  &43.50  &99.93  &\textbf{99.87}  &\textbf{99.98}  &97.65  &\textbf{98.60}  &94.05  \\ \hline
\multicolumn{1}{|l|}{Soft} &97.75  &99.84  &99.86  &99.91  &90.18  &\textbf{100}  &99.51  &98.68  &98.85  &98.37  &98.27  \\ \hline
\end{tabular}
\caption{Face forensic source classification accuracy reported on the proposed dataset.}
\label{table:ourdataset_hard_classify}
\end{table*}

\subsubsection{Face Forensic Localization}
We compare our methods with the aforementioned baselines for face forensic localization in \cref{table:ourdataset_localize} where average Intersection Over Union (IoU) values are reported. From the table, our methods achieve better accuracy than the proposed baselines with the best performance at 98.64\% by the soft version.
\begin{table*}[]
\centering
\begin{tabular}{c|c|c|c|c|c|c|c|c|c|c|c|}
\cline{2-12}
 & CelebA & FFHQ & DCGAN & LSGAN & BEGAN & WGAN & ProGAN & StyleGAN & StarGAN & SC-FEGAN & Total \\ \hline
\multicolumn{1}{|l|}{Encoder-Decoder} &98.93 & 98.37 & \textbf{100} & \textbf{100} & \textbf{100} & \textbf{100} & 98.94 & 98.78 & 97.37 & 94.39 & 98.06  \\ \hline

\multicolumn{1}{|l|}{Xception} &98.99  &98.82  &99.95  &\textbf{100}  &\textbf{100}  &\textbf{100}  &99.65  &98.69  &96.90  &95.05  &98.19  \\ \hline \hline

\multicolumn{1}{|l|}{Share} &99.41  &\textbf{99.99}  &\textbf{100}  &\textbf{100}  &\textbf{100}  &\textbf{100}  &\textbf{99.94}  &99.80  &97.23  &95.76  &98.62  \\ \hline

\multicolumn{1}{|l|}{Hard} &\textbf{99.97}  &90.94  &99.96  &\textbf{100}  &99.92  &\textbf{100}  &99.89  &\textbf{99.99}  &\textbf{97.59}  &96.51  &98.44  \\ \hline

\multicolumn{1}{|l|}{Soft} &99.80  &99.93  &\textbf{100}  &\textbf{100}  &\textbf{100}  &\textbf{100}  &99.86  &99.75  &97.29  &\textbf{97.08}  &\textbf{98.64}  \\ \hline
\end{tabular}
\caption{Localizatio accuracy reported on the proposed dataset.}
\label{table:ourdataset_localize}
\end{table*}

\subsection{FaceForecsic++ dataset}
The FaceForecsic++ is a video dataset collected from YouTube which contains pristine and edited videos based on three automatic facial manipulations methods: DeepFakes \cite{deepfakes}, Face2Face \cite{face2face}, and FaceSwap \cite{faceswap}. The dataset consist of 1,936,420 individual video frames at three compression rate: 0, 23, and 40.
\subsubsection{Face Forensic Detections}
The face forensic detection results are shown in \cref{table:ff_binary} where we compare our method with the methods reported in \cite{faceforensic++}. From the table, we can see that our model can achieve competitive results on high quality videos with accuracies 96.58\% and 94.85\% for videos at 0 and 23 compression rate respectively. This suggest that when the image quality is high, adding spatial landmarks information may disturb the signal directly coming from image pixels leading to a slight drop on the performance. On the other hand, when the image quality are low, facial landmarks play more important role. Notably, our method outperforms other methods significantly with 89.33\% in accuracy followed by a XceptionNet which achieve 85.49\%. 
\begin{table}[]
\centering
\begin{tabular}{|l|c|c|c|}
\hline
Methods \textbackslash Compressions & {Raw} & {HQ} & {LQ} \\ 
\hline
{Steg. features} & 97.63 & 70.78 & 56.37 \\ \hline
{Cozzolino \emph{et al.}} & 98.56 & 79.56 & 56.38 \\ \hline
{Bayar and Stamm} & 99.19 & 89.90 &70.01 \\ \hline
{Rahmouni \emph{et al.}} & 97.72 & 84.32 & 62.82 \\ \hline
{MesoNet} & 96.51 & 85.51 & 75.65 \\ \hline
{Encoder-Decoder} &90.28  &89.56  &85.13  \\ \hline
{XceptionNet}& \textbf{99.41} & \textbf{97.53} & 85.49 \\ \hline \hline
{Our} &96.58  &94.85  &\textbf{89.33}  \\ \hline
{Our-Mask} &96.21  &94.83  &89.29  \\ \hline
\end{tabular}
\caption{Binary detection accuracy reported on the FaceForensic++ dataset.}
\label{table:ff_binary}
\end{table}

\subsubsection{Face Forensic Localizations}
We also compare face forensic localization results with baselines methods on the FaceForecsic++ dataset in \cref{table:ff_local}. 
Similar to the results reported on face detection problems, our method performs slightly worse than the state-of-the-art on high-quality videos. Nevertheless, our method outperforms the baselines on low-quality videos achieving the IOU of 90.82\% whereas a method based on \cite{faceforensic++} only reach 90.40\%.

\begin{table}[]
\centering
\begin{tabular}{|l|c|c|c|}
\hline
Methods \textbackslash Compressions & {Raw} & {HQ} & {LQ} \\ 
\hline
\multicolumn{1}{|l|}{Encoder-Decoder} &91.59  &91.19  &86.68  \\ \hline
\multicolumn{1}{|l|}{XceptionNet} &\textbf{96.82}  &\textbf{95.53}  &90.40  \\ \hline \hline
\multicolumn{1}{|l|}{Our} &{96.72}  &{95.23}  &\textbf{90.82}  \\ \hline
\end{tabular}
\caption{Localization accuracy reported on the the FaceForensic++ dataset.}
\label{table:ff_local}
\end{table}

\subsection{Ablation study}
We conduct ablation study by comparing our methods in \cref{table:ourdataset_ablation}. We report the accuracies on face forensic binary detection (FBD), face forensic binary detection from a mask prediction (FBDM), face forensic type classification (FTC), face forensic source classification (FSC), and face forensic localization (FL). Firstly, we compare our method when class and mask branches are trained separately. From the table, we can see that apart from face forensic source classification (FSC), our multitask learning performs better for all other tasks. We also compare the benefit of adding landmarks information. The table demonstrated that spatial features from facial landmarks consistently improve the performance across all measurements. 

\begin{table}[]
\centering
\begin{tabular}{l|c|c|c|c|c|}
\cline{2-6}
\multicolumn{1}{c|}{} & FBD & FBDM & FTC & FSC & \multicolumn{1}{c|}{LC} \\ \hline
\multicolumn{1}{|l|}{Only class branch} &98.63 &- &98.51 &\textbf{98.42} &- \\ \hline
\multicolumn{1}{|l|}{Only mask branch} &- &99.13 &- &- &98.29 \\ \hline
\multicolumn{1}{|l|}{\textbf{Our}} &\textbf{99.20} &\textbf{99.25} &\textbf{98.67} &98.27 &\textbf{98.64} \\ \hline
\multicolumn{1}{|l|}{Our no landmarks} &98.88 &99.07 &98.27 &98.16 &98.25 \\ \hline
\end{tabular}
\caption{Ablation study reported on our proposed dataset.}
\label{table:ourdataset_ablation}
\end{table}

\subsection{Qualitative results}
\begin{figure*}
\includegraphics[width=\textwidth]{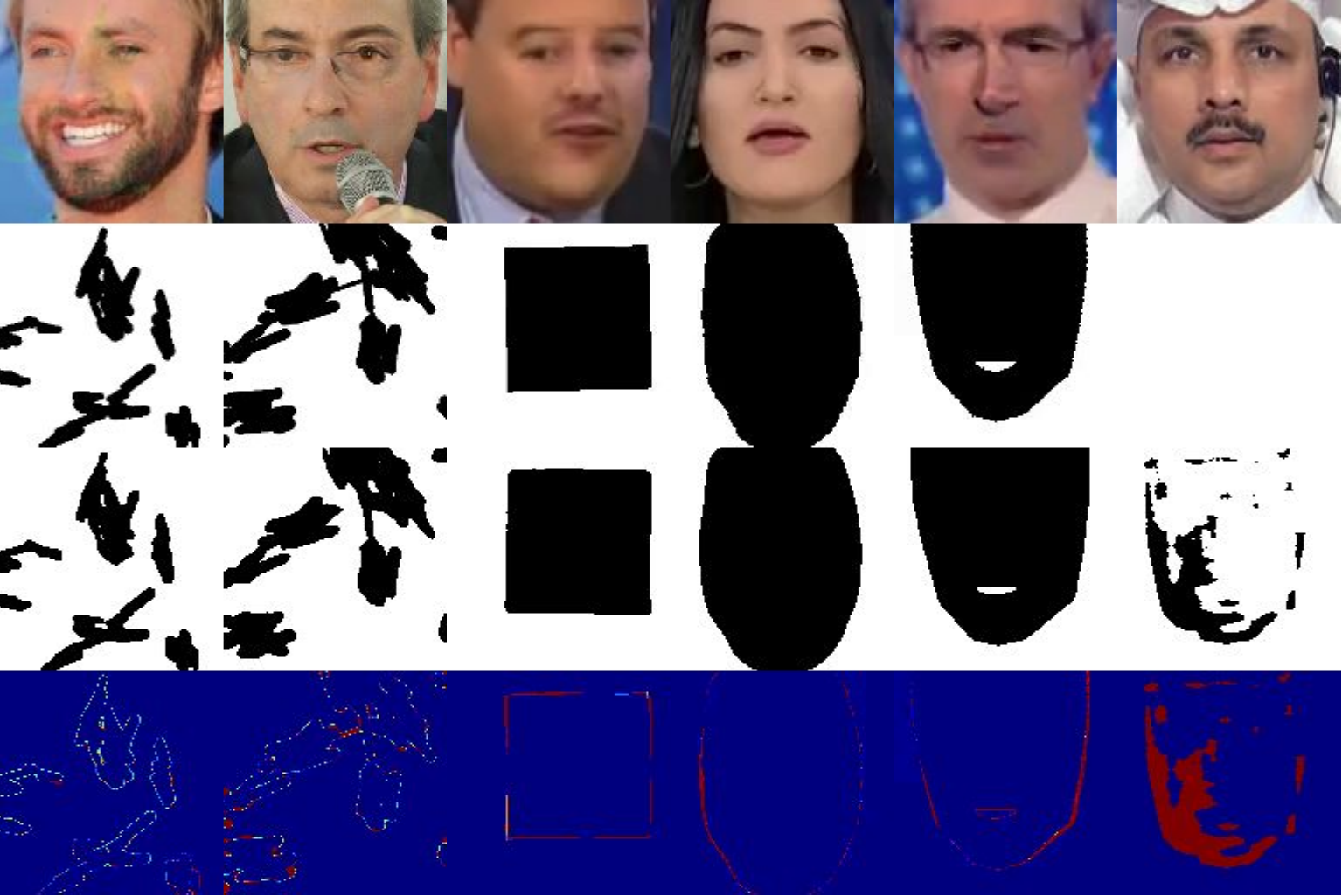}
\captionsetup[subfigure]{aboveskip=-1pt,belowskip=-1pt}
\begin{minipage}[t]{.164\linewidth}
\vspace{-0.8\baselineskip}
\subcaption{\scriptsize StarGAN}
\end{minipage}
\begin{minipage}[t]{.164\linewidth}
\vspace{-0.8\baselineskip}
\subcaption{\scriptsize SCFEGAN}
\end{minipage}\begin{minipage}[t]{.164\linewidth}
\vspace{-0.8\baselineskip}
\subcaption{\scriptsize DeepFake}
\end{minipage}\begin{minipage}[t]{.164\linewidth}
\vspace{-0.8\baselineskip}
\subcaption{\scriptsize Face2Face}
\end{minipage}
\begin{minipage}[t]{.164\linewidth}
\vspace{-0.8\baselineskip}
\subcaption{\scriptsize FaceSwap}
\end{minipage}
\begin{minipage}[t]{.164\linewidth}
\vspace{-0.8\baselineskip}
\subcaption{\scriptsize Real}
\end{minipage}
\caption{\label{fig:qualitative} Qualitative results on our dataset and the FaceForensic++ dataset. The first row is the given input images. The second row shows the ground-truth binary mask, and the third row show our network prediction. The last row show differences between the ground-truths and the predictions.}
\end{figure*}

In order to better understand the reported quantitative results, we have also shown qualitative results in \cref{fig:qualitative} with input images in the first row. The second and the third rows are ground-truth and the predicted masks produced by our method. The last row shows the heat map between the ground truth masks and the predicted masks. 
The figure demonstrates that our method can accurately localize manipulated locations. The last column shows some failure case where the method is ambiguous that the real image is an image edited by a Face2Face method.

\section{CONCLUSIONS AND FUTURE WORKS}

\subsection{Conclusions}
We propose to solve an important aspect of face forensic by introducing a large dataset for face forensic localization. We then proposed a model that exploits transfer learning, spatial facial landmarks, as well as combining prediction from class and mask predictions. Our method is a strong baseline for our dataset while also outperforms state-of-the-arts on the FaceForensic++ dataset.
\subsection{Future Works}
An interesting question regarding solving face forensic localization is whether a generator in GANs will benefit from a discriminator that can actually perform image localization rather than image classification. In particular, by properly including partially edited images subset during the training procedure, the discriminator should be able to not only classify an input image but also localize part of the image that needs to be improved by the generator.

% %%%%%%%%%%%%%%%%%%%%%%%%%%%%%%%%%%%%%%%%%%%%%%%%%%%%%%%%%%%%%%%%%%%%%%%%%%%%%%%%
% \section{ACKNOWLEDGMENTS}
% K. Songsri-in was supported by the Royal Thai Government Scholarship. The work of S. Zafeiriou was partially funded by EPSRC Project EP/N007743/1 (FACER2VM) and the FiDiPro Program of Tekes under Project 1849/31/2015.

% %%%%%%%%%%%%%%%%%%%%%%%%%%%%%%%%%%%%%%%%%%%%%%%%%%%%%%%%%%%%%%%%%%%%%%%%%%%%%%%%

\bibliographystyle{abbrv}
\bibliography{sample_FG2020}

\end{document}